\documentclass[10pt,twocolumn,letterpaper]{article}
\usepackage[pagenumbers]{cvpr}
\usepackage{amsthm}
\usepackage{multirow}
\usepackage{algorithm}
\usepackage{algpseudocode}
\usepackage{tikz}
\usetikzlibrary{arrows.meta,positioning,calc,fit,backgrounds}
\usepackage{cuted}
\usepackage{microtype}

\newcommand{\Fone}{F_{1}}
\newcommand{\Ftwo}{F_{2}}
\newcommand{\flow}{f}

\makeatletter
\renewcommand\paragraph{\@startsection{paragraph}{4}{\z@}%
  {-4pt}{1pt}{\normalfont\normalsize\bfseries}}
\let\sage@paragraph\paragraph
\renewcommand\paragraph[1]{%
  \sage@paragraph{#1}\nobreak\@afterindentfalse\@afterheading}
\makeatother

\let\oldbibliography\thebibliography
\renewcommand{\thebibliography}[1]{\oldbibliography{#1}\setlength{\itemsep}{0pt}\setlength{\parsep}{0pt}%
  \setlength{\parskip}{0pt}\setlength{\bibsep}{0pt}\fontsize{7.8}{9.1}\selectfont}

\definecolor{cvprblue}{rgb}{0.21,0.49,0.74}
\usepackage[pagebackref,breaklinks,colorlinks,allcolors=cvprblue]{hyperref}
\def\confName{CVPR}
\def\confYear{2027}
\title{Mapping the Design Space of Feature Sampling for Optical Flow}
\author{Liping Zhang}
\begin{document}
\maketitle
% Official author-kit order: the optional two-column teaser precedes the abstract.
% Full-width opening figure: qualitative comparison against WAFT.
% Source image: fig/qual_sintel.png (Sintel test clean, Ambush 1).
\begin{strip}
\centering
\includegraphics[width=0.94\textwidth]{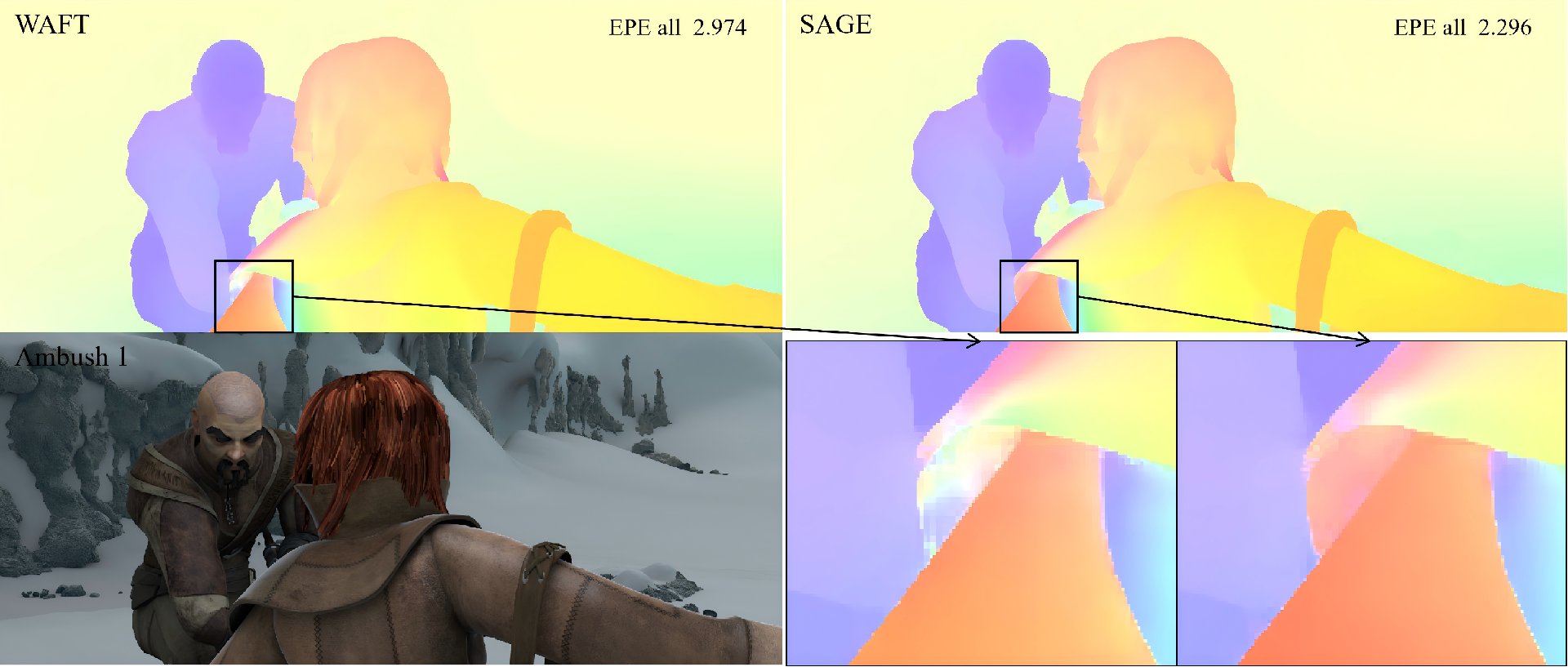}
\vspace{-1pt}
\captionof{figure}{\textbf{SAGE outperforms WAFT by 9.57\% in EPE-all on Sintel test clean, with a 16.2\% improvement on s40+ (4.627 vs.\ 5.521).} The figure presents a qualitative comparison on the Ambush 1 scene from Sintel test clean, where SAGE produces more accurate outputs with sharper boundaries.}
\label{fig:teaser}
\vspace{4pt}
\end{strip}

\begin{abstract}
The computation of correlation features is crucial for optical flow estimation. Existing methods (e.g., RAFT) perform fixed-radius lookups on a cost volume, yielding a set of correlation features for each position; alternatively, WAFT performs single-point feature indexing, yielding one correlation feature vector per position. None of them designs a suitable correlation feature sampling scheme for each position according to regional characteristics. Our insight is that not every position requires the same sampling scheme. Applying an identical sampling scheme to every position is either redundant or insufficient and ignores the specific characteristics of the corresponding region. This paper rethinks the design of feature sampling and investigates the following question: should the support of feature sampling---i.e., the extent and directions over which evidence is aggregated---be learned separately for each query position? To study this question, we propose SAGE (Sampling with Anisotropic Gaussian Expectation). This method replaces single-point sampling with a per-pixel anisotropic Gaussian expectation on a fixed candidate grid centered on the current optical flow estimate. A positive-definite precision matrix controls the scale, anisotropy, and orientation of the sampling support, with single-point sampling included as its limiting case. Consequently, the sampling support becomes an explicit and analyzable design variable and provides a principled way to analyze how directional feature aggregation affects correspondence learning. Extensive experiments show that SAGE, while adding only 1.5\% multiply-accumulate operations (MACs), effectively improves zero-shot cross-dataset generalization. When trained only on FlyingChairs and FlyingThings, SAGE reduces the end-point error (EPE) by 24.2\% and 16.4\% on Sintel clean and final, and reduces KITTI F1-epe and F1-all by 7.2\% and 11.6\%. The code will be made publicly available at \url{https://github.com/LipingZhang17/SAGE}.
\end{abstract}

\section{Introduction}
\label{sec:intro}

Optical flow estimation requires retrieving correlation feature information from two adjacent input frames. That is, it must answer the question of where and how to obtain such correlation features. Feature sampling based on a single location retrieves correlation features from only one spatial location at each query position, and therefore cannot explicitly control the spatial support of the sampled features. We revisit this design choice and propose the following insight: the sampling of correlation features at each position should be determined in a learned rather than handcrafted manner, according to the characteristics and requirements of that position, including the sampling range, direction, and number of correlation features. Specifically, for positions with clear and unambiguous correspondences, such as unoccluded small-displacement regions, retrieving features from a single location is sufficient and reliable. In contrast, for positions with ambiguous and uncertain matching, such as occlusions, large displacements, motion boundaries, and weak-texture regions, a single correlation feature vector is insufficient and unreliable. This poses severe challenges to iterative refinement modules, which generally struggle to reverse erroneous estimates. Therefore, it is necessary to design a learnable sampling scheme for each position according to the difficulty of the problem it must handle. Such a scheme should account for the spatial extent, direction, and weighting of the correlation features required at each position.

In this paper, we propose SAGE (Sampling with Anisotropic Gaussian Expectation), which places normalized Gaussian weights on a fixed $K{\times}K$ grid centered at the current flow estimate. Its per-pixel positive-definite precision matrix is jointly predicted from the source features, the warped target features, and their mismatch. As the precision tends to infinity, SAGE exactly degenerates to single-point sampling; therefore, the comparison with the baseline operator is a limiting case rather than a comparison with another model. Isotropic averaging corresponds to the scalar-precision case, allowing the shape to be varied independently at a fixed width. The precision carries three interpretable degrees of freedom---width, eigenvalue ratio, and orientation---so changes in behavior can be attributed to a named geometric property rather than to additional parameters. Under a single training recipe and zero-shot transfer setting, SAGE improves cross-dataset performance over WAFT, reducing Sintel clean and final EPE by $24.2\%$ and $16.4\%$, and KITTI $\Fone$-epe and $\Fone$-all by $7.2\%$ and $11.6\%$. \Cref{fig:teaser} visualises the comparison against WAFT on Sintel test clean, where SAGE performs better in fast-motion regions.

\paragraph{Contributions.} (i)~We reformulate the design of correlation feature sampling, propose a new perspective for feature sampling, and cast it as a four-axis design space---signal, center, adaptive granularity, and weighting shape. (ii)~We propose SAGE as a differentiable member of this correlation feature sampling space; both single-point sampling and isotropic sampling are exact limiting cases of it, making the sampling shape a controllable variable. (iii)~SAGE effectively improves the cross-dataset generalization performance of state-of-the-art optical flow estimators, with gains mainly concentrated in occlusion, large-displacement, motion-boundary, and weak-texture regions.

\section{Related Work}
\label{sec:related}

\textbf{Explicit local correlation sampling.} FlowNetC~\cite{flownet2} introduces a correlation layer. In effect, this operation can be regarded as extracting, for each $\mathbf{x}_1$, a local discrete displacement slice centred at $\mathbf{x}_2=\mathbf{x}_1$ (zero displacement) from the all-pairs 4D cost volume. PWC-Net~\cite{pwcnet} designs the sampling of correlation features to be performed at each level of a feature pyramid, progressively amplifying the effect of this hand-crafted local search window through ``pyramid $+$ backward warping'', thereby covering large displacements. RAFT~\cite{raft} and SEA-RAFT~\cite{sea_raft} instead construct a multi-scale all-pairs 4D cost volume (correlation pyramid). During iterative updates, they use the current optical flow to forward-warp $\mathbf{x}_1$, obtain its floating-point corresponding position $\mathbf{x}_1+\flow_{1\to2}(\mathbf{x}_1)$ in the second frame, and, centred at this position, perform bilinear interpolation lookup on the cost volume over a fixed local offset grid to obtain sparse local correlation features for updating the optical flow. GMA~\cite{gma} keeps the local sampling unchanged on the basis of RAFT's local lookup and additionally adds a ``global self-attention aggregation'' pathway, explicitly transferring motion information from unoccluded regions to occluded regions. Ef-RAFT~\cite{efraft} proposes an amorphous lookup operator (ALO), which reduces the data redundancy of RAFT's search operator and enlarges the search space for similarity extraction, so that the search operator is no longer restricted to a fixed local square window. DCVNet~\cite{dcvnet} constructs multiple dilated cost volumes with different dilation factors to capture both small and large displacements simultaneously, forming a structured multi-scale sampling. SKFlow~\cite{skflow}, in turn, uses large-kernel convolution and a selective kernel mechanism to turn the sampling scale into a process dynamically selected by the network according to content, allowing the network to decide by itself at which scale to perform matching.

\begin{table*}[t]
\centering\footnotesize\setlength{\tabcolsep}{6.0pt}
\begin{tabular}{@{}l l l l l@{}}
\toprule
Read-out & Signal & Centre & Granularity & Weight $A[p,j]$ of candidate $j$ \\
\midrule
RAFT~\cite{raft} & similarity & $\flow_t[p]$ & fixed grid & $1/K^2$ \\
Ef-RAFT~ALO~\cite{efraft} & similarity & $s_\ell\Delta_j+\mathbf{t}_\ell$ & scale and shift per level & $1/K^2$ \\
WAFT~\cite{waft} & feature & $\flow_t[p]$ & none & $\mathbf{1}[\Delta_j{=}0]$ \\
SAGE-Iso (ours) & feature & $\flow_t[p]$ & per pixel, scalar precision & $\propto\exp(-\lVert\Delta_j\rVert^2/2b[p]^2)$ \\
SAGE (ours) & feature & $\flow_t[p]$ & per pixel, full precision & $\propto\exp(-\tfrac12\Delta_j^\top\Sigma^{-1}[p]\Delta_j)$ \\
\bottomrule
\end{tabular}
\caption{Feature read-outs of prior work and of this paper; symbols are defined in \cref{sec:design}. The last column gives the unnormalised candidate weight; weights are normalised over the $K^2$ candidates, and $\propto$ denotes proportionality. The first three rows are not retrained here: their entries state the published form, and they are listed so that the remaining rows can be compared against them.}
\label{tab:readouts}
\end{table*}

\textbf{Global/attention matching as an alternative to local sampling.} CRAFT~\cite{craft} abandons the explicit local sampling window and replaces pairwise dot-product correlation with cross-frame attention: features from one frame are used as Query and features from the other frame as Key/Value, and the global matching cost is computed via an attention mechanism, at the cost of a significant increase in computation. FlowFormer~\cite{flowformer,flowformer_pp} tokenises the $H\times W\times H\times W$ volume into cost tokens and encodes them with alternating grouped Transformer layers, switching between intra-map and inter-map attention, so global matching evidence is aggregated into a cost memory at controlled complexity. GMFlow~\cite{gmflow} and GMFlowNet~\cite{gmflownet} instead directly apply an $\operatorname{argmax}$ operation on the 4D cost volume to compute global matching, finding the globally optimal corresponding point for each pixel.

\textbf{Sparse matching and cost-volume-free sampling paradigms.} SparseFlow~\cite{sparseflow} does not perform dense matching directly; instead, it first extracts sparse pixel-level correspondences by solving a sparse reconstruction optimisation problem to select reliable matches in a data-driven manner, and then densifies them with a variational method. WAFT~\cite{waft} argues that a cost volume is not a necessary condition for strong performance; on high-resolution feature maps at $1/2$ resolution, it backward-warps the second-frame features using the current optical flow estimate and extracts feature vectors at the corresponding positions via bilinear interpolation. 

% 通栏插图：同场景下 P(x,y)、单点采样（WAFT）与 SAGE 的读出对比。
% 图片来源：fig/readout_compare.png。
% 用 table*[t] 置于页顶（cuted 的 strip 在此处会把图放到页底，且上方留下整块空白）。
\begin{figure*}[!t]
\centering
\vspace*{-2pt}
\includegraphics[width=0.9\textwidth]{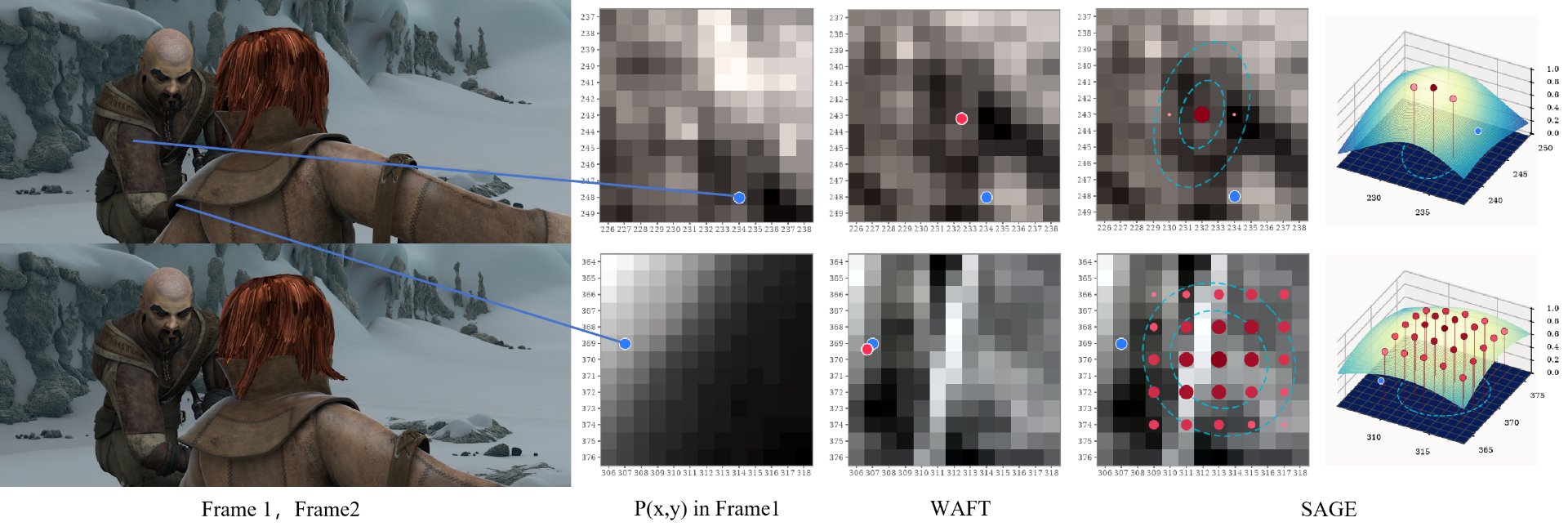}
\vspace*{-4pt}
\caption{Taking a pair of frames from Sintel test clean as an example, SAGE samples the corresponding correlation features for positions in different regions. In small-displacement regions the sampling of SAGE is essentially the same as that of WAFT; in fast-motion unmatched regions, SAGE samples richer features than the single nearby feature of WAFT and is therefore more robust there. The resulting flows are visualised in \cref{fig:teaser}.\label{fig:readout}}
\vspace*{4pt}
\end{figure*}

\section{Method}
\label{sec:method}

\subsection{Precision prediction and expected feature}
\label{sec:design}

SAGE has two parts: a head that predicts a per-pixel weight map, and an expectation that turns that weight map into a feature. The discussion and experiments below take the single-point sample of WAFT as the baseline.

\begin{figure}[!t]
\centering
\includegraphics[width=\columnwidth]{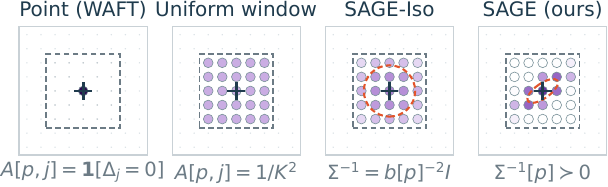}
\caption{\textbf{The read-out as a one-parameter family.} The four read-outs of \cref{tab:readouts} on one shared $9\times9$ offset lattice, the dashed frame marking the central $5\times5$ candidate region over which the weights are normalised. All four panels read the same $5\times5$ candidate grid; they differ only in how the weight is placed on it. Disc colour encodes $A[p,j]$ on a common logarithmic scale, and the cross marks the current flow estimate $q_0=p+\flow_t[p]$; grey dots are zero-weight candidates and the warm dashed curve is the equal-logit contour.}
\label{fig:lattice}
\end{figure}

\paragraph{Precision prediction.} The per-pixel weight prediction module consists of two $1\times1$ convolutions with a ReLU between them. Its input concatenates the source feature, the warped target feature and the difference between them
\begin{equation}
 x[p]=\big[\Fone[p],\;\Ftwo^{\mathrm{warp}}[p],\;\big|\Fone[p]-\Ftwo^{\mathrm{warp}}[p]\big|\big]\in\mathbb{R}^{3D}
 \label{eq:head_in}
\end{equation}
where the warped target feature is
\begin{equation}
 \Ftwo^{\mathrm{warp}}[p]=\Ftwo\big[p+\flow_t[p]\big]
 \label{eq:warp}
\end{equation}
that is, the feature the single-point sample would return. The layer sizes of the head are given in \cref{tab:head}: for $D=64$ it maps a $192\times H\times W$ tensor to $3\times H\times W$.

\begin{table}[h!]
\centering\footnotesize\setlength{\tabcolsep}{4.5pt}
\begin{tabular}{@{}lcc@{}}
\toprule
Operation & Input size & Output size \\
\midrule
$1\times1$ conv $+$ ReLU & $192\times H\times W$ & $64\times H\times W$ \\
$1\times1$ conv & $64\times H\times W$ & $3\times H\times W$ \\
\bottomrule
\end{tabular}
\caption{Layer sizes of the weight prediction head. With $D=64$ the input has $3D=192$ channels and the output three, which are the three non-zero entries of $L[p]$ in \cref{eq:chol}.}
\label{tab:head}
\end{table}

The three output channels fill a lower-triangular factor by column
\begin{equation}
 L[p]=
 \begin{pmatrix}
 \ell_{11}[p] & 0\\[2pt]
 \ell_{21}[p] & \ell_{22}[p]
 \end{pmatrix}
 \label{eq:chol}
\end{equation}
from which the precision follows as
\begin{equation}
 \Sigma^{-1}[p]=L[p]^\top L[p]
 \label{eq:precision}
\end{equation}
with $\ell_{11}$ and $\ell_{22}$ obtained by exponentiating their channels and applying a floor, so $L[p]$ is non-singular and $\Sigma^{-1}[p]$ positive definite by construction. Intuitively, $L[p]$ is a square root of the precision: predicting $L$ rather than $\Sigma^{-1}$ makes positive definiteness a structural property rather than something the objective has to maintain, and no matrix inverse is ever computed in the forward pass. The factorisation is a parameterisation, not a computation---the three free entries span the whole cone of symmetric positive definite $2\times2$ matrices, with no constraint beyond positive definiteness.

Candidates are drawn from a $K\times K$ grid in the second frame centred at $q_0=p+\flow_t[p]$; their offsets are written $\Delta_j$, with $\Delta_0=(0,0)$ at the current warp. The logit of a candidate is the squared length of its offset after the linear map $L[p]$
\begin{equation}
 z[p,j]=-\tfrac12\big\|L[p]\,\Delta_j\big\|_2^2
 \label{eq:logit}
\end{equation}
and the weights are these logits in a softmax over the grid
\begin{equation}
 A[p,j]=\frac{\exp\big(z[p,j]\big)}{\sum_{k=0}^{K^2-1}\exp\big(z[p,k]\big)}
 \label{eq:weights}
\end{equation}
Intuitively, $z$ is a distance measured in units of local precision, so a candidate is discounted in proportion to how far it departs along the direction in which the pixel is uncertain. Since $\Delta_0=(0,0)$ gives $z[p,0]=0$, the weights depend only on the offset and on $L[p]$: every candidate on the fixed grid receives a normalised weight, and any position outside the grid receives none. Each candidate costs two multiply-adds.

\paragraph{Against the single-point sample.} WAFT reads one feature at $q_0$, and only that one
\begin{equation}
 \Ftwo^{\mathrm{pt}}[p]=\Ftwo[q_0]
 \label{eq:ptread}
\end{equation}
with a one-hot selection weight
\begin{equation}
 A^{\mathrm{point}}[p,j]=\mathbf{1}\big[\Delta_j=0\big]
 \label{eq:ptweight}
\end{equation}
The sampling design we rebuild keeps the same centre $q_0$ and replaces the single sample at that point by a weighted average over the candidate grid\label{sec:centre}

\begin{equation}
 \Ftwo^{\mathrm{SAGE}}[p]=\sum_{j}A[p,j]\,\Ftwo\big[p+\flow_t[p]+\Delta_j\big]
 \label{eq:sage}
\end{equation}
with weights $A[p,j]$ from \cref{eq:weights}. The two therefore differ only in what is read and in how it is read. \Cref{fig:readout} shows the resulting sampling pattern of SAGE and WAFT at two query positions of one instance.

%=========================================================
\subsection{The operator as a nested family}
\label{sec:family}

Equation~\eqref{eq:weights} describes the read-out as a distribution over the grid, so the free object is the weight map $A[p,\cdot]$. Two restrictions clarify what the head can and cannot express, and they are the reason the family is worth parameterising rather than simply predicting weights.

\paragraph{Isotropic restriction.} Setting $\ell_{21}=0$ and $\ell_{11}=\ell_{22}=b[p]^{-1}$ gives the scalar precision $\Sigma^{-1}[p]=b[p]^{-2}I$ and the weights
\begin{equation}
 A^{\mathrm{iso}}[p,j]\propto\exp\!\Big(-\frac{\lVert\Delta_j\rVert_2^2}{2\,b[p]^2}\Big)
 \label{eq:iso}
\end{equation}
which is the SAGE-Iso row of \cref{tab:readouts}. The head then emits one channel, and only the width $b[p]$ of a circularly symmetric support adapts per pixel.

\paragraph{Arbitrary distribution.} Predicting $K^2-1$ free logits gives an arbitrary categorical distribution. It is more expressive but costs the property that makes this study possible: its entries carry no geometric meaning, so a change in accuracy could not be assigned to a named cause. The quadratic form is the smallest restriction under which the learned quantity stays readable (\cref{sec:centre,sec:conditioning}).

\paragraph{Limiting cases.} Two limits fix the position of the operator. If all entries of $L[p]$ grow without bound, then $z[p,j]\to-\infty$ for every $j\neq0$ while $z[p,0]=0$, hence
\begin{equation}
 A[p,\cdot]\longrightarrow A^{\mathrm{point}}[p,\cdot]
 \label{eq:limitpoint}
\end{equation}
and Eq.~\eqref{eq:sage} returns the single-point sample of Eq.~\eqref{eq:ptread} exactly. If instead all entries of $L[p]$ tend to zero, all logits vanish and
\begin{equation}
 A[p,\cdot]\longrightarrow \mathbf{1}/K^2
 \label{eq:limituniform}
\end{equation}
the uniform average over the grid. These are limits of the parameterisation, not descriptions of the trained operator: the learned weights lie away from both ends, and the distance to either end is what the grid size controls (\cref{sec:ablations}). WAFT sampling is therefore not a different architecture but one end of one axis, so \cref{sec:main} reports movement along that axis rather than a contest between designs.

%=========================================================
\subsection{Geometry, conditioning, and scale invariance}
\label{sec:conditioning}

Write the precision matrix in general form as
\begin{equation}
 \Sigma^{-1}[p]=\begin{pmatrix}a[p] & b[p]\\ b[p] & c[p]\end{pmatrix}.
 \label{eq:prec2x2}
\end{equation}
Its spectral decomposition
\begin{equation}
\begin{aligned}
 \Sigma^{-1}[p]&=\lambda_1\mathbf{v}_1\mathbf{v}_1^\top+\lambda_2\mathbf{v}_2\mathbf{v}_2^\top\\
 \lambda_1&\geq\lambda_2>0
\end{aligned}
 \label{eq:eig}
\end{equation}
gives two precisions and two orthonormal directions. The constant-logit contour is an ellipse with semi-axes proportional to $\lambda_1^{-1/2}$ and $\lambda_2^{-1/2}$; the eigenvector $\mathbf{v}_1$ of the larger precision points along the short axis, where weight decays fastest, and $\mathbf{v}_2$ along the long axis, along which the read-out averages. Because $\Sigma^{-1}[p]$ is a full matrix, the read-out can average along one direction while staying sharp along another---the behaviour a motion boundary requires, since averaging along an edge is harmless whereas averaging across it mixes two independent motions.

The logit of Eq.~\eqref{eq:logit} makes the conditioning explicit: $\Delta_j$ is mapped by $L[p]$ into a space where distance is measured in units of local precision, so the channels producing $\ell_{11}$ and $\ell_{22}$ set the decay along the principal axes and the channel producing $\ell_{21}$ sets their relative orientation. Since the head also receives the mismatch $|\Fone-\Ftwo^{\mathrm{warp}}|$, the precision is conditioned on how well the estimate aligns the frames, which an appearance-only head could not establish.

One normalisation property holds by construction: the weights are invariant to a common rescaling of $L[p]$, since a positive constant multiplies every logit and leaves the softmax unchanged. Only the eigenvalue ratio and the orientation are identifiable from $A[p,\cdot]$, not the magnitude of the precision, which we summarise by an effective scalar width
\begin{equation}
 \sigma_{\mathrm{eff}}[p]=\det\big(\Sigma^{-1}[p]\big)^{-1/4}
 \label{eq:sigmahat}
\end{equation}
Intuitively, $\sigma_{\mathrm{eff}}$ is the radius of the circular Gaussian that would enclose the same area, so read-outs of different shape can be placed on one axis; it equals $b[p]$ under Eq.~\eqref{eq:iso}. Equation~\eqref{eq:sigmahat} is used below only as that geometric summary and is not read as a calibrated posterior over correspondence.

%=========================================================
\subsection{Computational cost}
\label{sec:cost}

The expectation of Eq.~\eqref{eq:sage} samples $K^2$ candidates and forms their weighted sum, at a cost of $C_{\mathrm{samp}}=HWK^2D$ multiply--accumulates, to which Eq.~\eqref{eq:logit} adds two per candidate; the head of Eq.~\eqref{eq:head_in} applies two $1\times1$ convolutions of sizes $3D\times D$ and $D\times c_{\mathrm{out}}$, at a cost of
\begin{equation}
 C_{\mathrm{head}}=HW\big(3D^2+Dc_{\mathrm{out}}\big)
 \label{eq:costhead}
\end{equation}
with $c_{\mathrm{out}}=3$ for SAGE and $c_{\mathrm{out}}=1$ for SAGE-Iso. At $K=5$, $D=64$, and $HW=128^2$, sampling adds about $0.003\%$ of the $807$ GMACs of the baseline whereas the head adds about $1.5\%$; both figures are quoted because the grid term alone would understate the cost by two orders of magnitude. SAGE-Iso pays the same sampling cost and differs only in its output layer, which at $D=64$ amounts to $2D+2=130$ parameters. No matrix inverse and no change to the encoder is involved.

These figures carry an argument. Read-out shape is available for roughly one part in seventy of the compute budget, far below the cost of changing an encoder. If it nevertheless changes cross-dataset transfer by double-digit percentages, then its role cannot be that of a rounding term.

\section{Experiments}
\label{sec:exp}

\subsection{Protocol and evaluation record}
\label{sec:setup}

All zero-shot experiments follow the training configuration of WAFT, given in \cref{tab:train-recipe}; We take WAFT-DINOv3-a2 as the baseline model~\cite{dinov3} in this paper. Zero-shot evaluation is on the Sintel and KITTI~2015 training splits and Spring sub-val~\cite{sintel,kitti,spring}. For Sintel, we report EPE and the regional metrics (Matched, Unmatched, s0--10, s10--40, and s40+) on both the Clean and Final passes. For KITTI, we report Fl-all (\%) and Fl-epe (px). For Spring, we report EPE, 1-pixel error, Fl, and regional metrics of s0--10, s10--40 and s40+.

\begin{table}[t]
\centering\scriptsize\setlength{\tabcolsep}{2.6pt}
\begin{tabular}{@{}llcc@{}}
\toprule
Training data & Crop & Steps & LR \\
\midrule
\multicolumn{4}{@{}l}{\emph{Benchmark submissions}} \\
TartanAir & $480{\times}640$ & $300$k & $4{\times}10^{-4}$ \\
FlyingChairs & $368{\times}496$ & $50$k & $4{\times}10^{-4}$ \\
FlyingThings & $432{\times}960$ & $200$k & $4{\times}10^{-4}$ \\
KITTI train & $432{\times}1242$ & $5$k & $1{\times}10^{-4}$ \\
Things $+$ HD1K $+$ KITTI $+$ Sintel & $432{\times}960$ & $200$k & $1{\times}10^{-4}$ \\
Spring train ($540$p) & $540{\times}960$ & $200$k & $1{\times}10^{-4}$ \\
\midrule
\multicolumn{4}{@{}l}{\emph{Zero-shot evaluation}} \\
FlyingChairs & $368{\times}496$ & $50$k & $1{\times}10^{-4}$ \\
FlyingThings & $432{\times}960$ & $50$k & $1{\times}10^{-4}$ \\
\bottomrule
\end{tabular}
\caption{Training recipe for benchmark submissions (top) and zero-shot evaluation (bottom).}
\label{tab:train-recipe}
\end{table}

\begin{table}[t]
\centering\scriptsize\setlength{\tabcolsep}{2.6pt}%
\renewcommand{\arraystretch}{1.06}%
\begin{tabular}{@{}lcccc@{}}
\toprule
\multirow{2}{*}{Method} & \multicolumn{2}{c}{Sintel (train)} & \multicolumn{2}{c}{KITTI (train)} \\
\cmidrule(lr){2-3}\cmidrule(lr){4-5}
 & Clean$\downarrow$ & Final$\downarrow$ & Fl-epe$\downarrow$ & Fl-all$\downarrow$ \\
\midrule
PWC-Net~\cite{pwcnet} & 2.55 & 3.93 & 10.4 & 33.7 \\
RAFT~\cite{raft} & 1.43 & 2.71 & 5.04 & 17.4 \\
GMA~\cite{gma} & 1.30 & 2.74 & 4.69 & 17.1 \\
SKFlow~\cite{skflow} & 1.22 & 2.46 & 4.27 & 15.5 \\
DIP~\cite{dip} & 1.30 & 2.82 & 4.29 & 13.7 \\
EMD-L~\cite{emdl} & 0.88 & 2.55 & 4.12 & 13.5 \\
CRAFT~\cite{craft} & 1.27 & 2.79 & 4.88 & 17.5 \\
RPKNet~\cite{rpknet} & 1.12 & 2.45 & -- & 13.0 \\
GMFlowNet~\cite{gmflownet} & 1.14 & 2.71 & 4.24 & 15.4 \\
FlowFormer~\cite{flowformer} & 1.01 & 2.40 & 4.09 & 14.7 \\
FlowFormer++~\cite{flowformer_pp} & 0.90 & 2.30 & 3.93 & 14.2 \\
CCMR+~\cite{ccmr} & 0.98 & 2.36 & -- & 12.9 \\
MatchFlow(G)~\cite{matchflow} & 1.03 & 2.45 & 4.08 & 15.6 \\
SEA-RAFT(L)~\cite{sea_raft} & 1.19 & 4.11 & 3.62 & 12.9 \\
AnyFlow~\cite{anyflow} & 1.10 & 2.52 & 3.76 & 12.4 \\
SAMFlow~\cite{samflow} & \underline{0.87} & \textbf{2.11} & 3.44 & 12.3 \\
FlowDiffuser~\cite{flowdiffuser} & \textbf{0.86} & 2.19 & 3.61 & 11.8 \\
DPFlow~\cite{dpflow} & 1.02 & 2.26 & \underline{3.37} & \textbf{11.1} \\
\midrule
WAFT-DINOv3-a2~\cite{waft} & 1.28 & 2.56 & 3.49 & 12.9 \\
SAGE (ours) & 0.97 & \underline{2.14} & \textbf{3.24} & \underline{11.4} \\
\bottomrule
\end{tabular}
\caption{SAGE improves zero-shot cross-dataset generalisation, reducing EPE by $24.2\%$ and $16.4\%$ on Sintel clean and final, and Fl-epe and Fl-all by $7.2\%$ and $11.6\%$ on KITTI. Bold marks the best value and an underline the second best.}
\label{tab:waft-zeroshot}
\end{table}

Sintel and Spring use pixel-weighted micro aggregation; KITTI Fl-epe is frame-averaged. Evaluation runs at each benchmark's native resolution, and all models are evaluated under one deterministic inference setting; the exact setting and the three evidence sources are given in the supplementary material.

\subsection{Comparison of sampling methods}
\label{sec:main}
\begin{table}[t]
\centering\scriptsize
\setlength{\tabcolsep}{1.9pt}%
\renewcommand{\arraystretch}{0.94}%
\begin{tabular}{@{}lccccccc@{}}
\toprule
& \multicolumn{2}{c}{Sintel} & \multicolumn{2}{c}{KITTI} & \multicolumn{3}{c}{Spring} \\
\cmidrule(lr){2-3}\cmidrule(lr){4-5}\cmidrule(lr){6-8}
Method\hspace{2pt} & Clean$\downarrow$ & Final$\downarrow$ & Fl-all$\downarrow$ & Fl-epe$\downarrow$ & EPE$\downarrow$ & 1px$\downarrow$ & Fl$\downarrow$ \\[1.2pt]
\midrule
WAFT & 1.28 & 2.56 & 12.9 & 3.49 & \underline{0.18} & \textbf{1.02} & 0.43 \\
ALO-global & \underline{0.988} & 2.558 & 12.639 & 3.735 & 0.184 & 1.054 & \textbf{0.416} \\
ALO-pixel & 1.025 & \underline{2.415} & 12.148 & \underline{3.247} & 0.187 & 1.140 & 0.454 \\
SAGE-iso & 1.036 & 2.731 & \underline{11.929} & 3.678 & \textbf{0.182} & \underline{1.032} & 0.446 \\
SAGE & \textbf{0.965} & \textbf{2.141} & \textbf{11.366} & \textbf{3.237} & 0.185 & 1.317 & \underline{0.419} \\
\bottomrule
\end{tabular}
\caption{Evaluation of sampling  methods; SAGE attains the best cross-dataset performance.}
\label{tab:controlled}
\end{table}

For fair evaluation, we transfer the ALO sampling method to the baseline model, train from scratch and test its zero-shot cross-dataset transfer performance; the experiments cover a globally normalised ALO-global and a per-pixel ALO-pixel variant, and the results are listed in \cref{tab:controlled}. 
Both ALO-global and ALO-pixel improve model accuracy.Notably, ALO-pixel’s per-pixel query method outperforms the global setting that applies a unified query operator to all positions, providing experimental support for our claim that separate sampling methods should be learned for each position in the scene.
Experimental results from isotropic SAGE-iso and anisotropic SAGE further corroborate this observation.
SAGE-iso shares the grid, the centre and the signal with SAGE and only constrains the precision matrix to be isotropic, so the gap between the two isolates shape: $6.9\%$ on clean, $21.6\%$ on final, $4.7\%$ on Fl-all and $12.0\%$ on Fl-epe.
Although the isotropic design learns a feature sampling distribution for each position, it does not explicitly account for the orientation of image boundaries and motion boundaries in real scenes. Results show that, except for an improvement in F1-all on KITTI, performance degrades on all other datasets, indicating that the isotropic design captures redundant and interfering features and thereby introduces bias into iterative estimation.

SAGE achieves the best cross-dataset generalization on KITTI(train) and Sintel(train) acrose 5 sampling methods.On Spring sub-val set, SAGE improves Fl by $2.6\%$ and the s10--40 EPE by $4.4\%$, but raises the overall EPE by $1.1\%$ and the 1px error from $1.02$ to $1.317$.
Cross-dataset generalization evaluation of zero-shot models under different sampling methods further demonstrates that SAGE learns a weighted anisotropic sampling feature set for each position across the entire scene. Compared with WAFT, which samples only a single position, SAGE is more robust to large displacements and occluded regions. Compared with ALO-global and ALO-pixel, which expand the search range globally or per principal pixel, SAGE provides more reliable correlation feature inputs for iterative refinement. Compared with isotropic SAGE, SAGE incurs less feature sampling redundancy across different regions, making it an effective design for adapting to region-specific scene characteristics.

\subsection{Performance analysis}
\label{sec:shape}
SAGE achieves the best performance on 11 of all 12 regional metrics; in the s0–10 region of Spring sub-val set, its EPE is 0.004 higher than that
of WAFT, sees (Table~1 of the supplementary material). 
From the design principle of SAGE, compared with directly sampling a feature at a single position, the s0–10 region reflects the effect of SAGE’s weighted sampling: replacing a single sample with a weighted mean eliminates extreme values caused by sampling the wrong candidate, while introducing a small bias where the true correspondence is sharper than the kernel.When the single-point sample lands correctly, averaging only blurs; when it may land across a boundary, averaging recovers evidence that single-point sampling drops. SAGE-iso shows the complementary pattern---best at small displacements, worse than WAFT at medium ones($0.350\to0.340$)---because a circular kernel averages across boundaries while an elliptical one runs along them.

Results on Sintel reproduces the same ordering on a different error structure. The largest gains are exactly where single-point feature sampling is least reliable: unmatched $8.874\to7.649$ ($-13.8\%$) and large displacement s40+  clean:$7.569\to6.231$ ($-17.7\%$), final:$15.976\to14.454$ ($-9.53\%$) , both above the all-pixel gain of the same run ($13.1\%$). Occlusion and displacement are two independent partitions of the same predictions, SAGE improves performance by mitigating the unreliability of single-point feature sampling, thereby effectively enhancing the model’s generalization in unmatched and large-displacement regions.

KITTI was collected on real roads and contains real-world challenges including large displacements, low/repetitive texture, non-Lambertian surfaces, and severe illumination changes.SAGE achieves better generalization performance(Fl-epe and Fl-all by $7.2\%$ and $11.6\%$) on such real-world scene datasets. Compared with sampling a single correlation feature for positions across all regions, SAGE benefits from acquiring richer correlation features for regions with large displacements and weak textures, where a single feature vector is weakly discriminative regardless of displacement thereby improving the robustness of the method.

\begin{figure*}[!t]
\centering
\includegraphics[width=\textwidth]{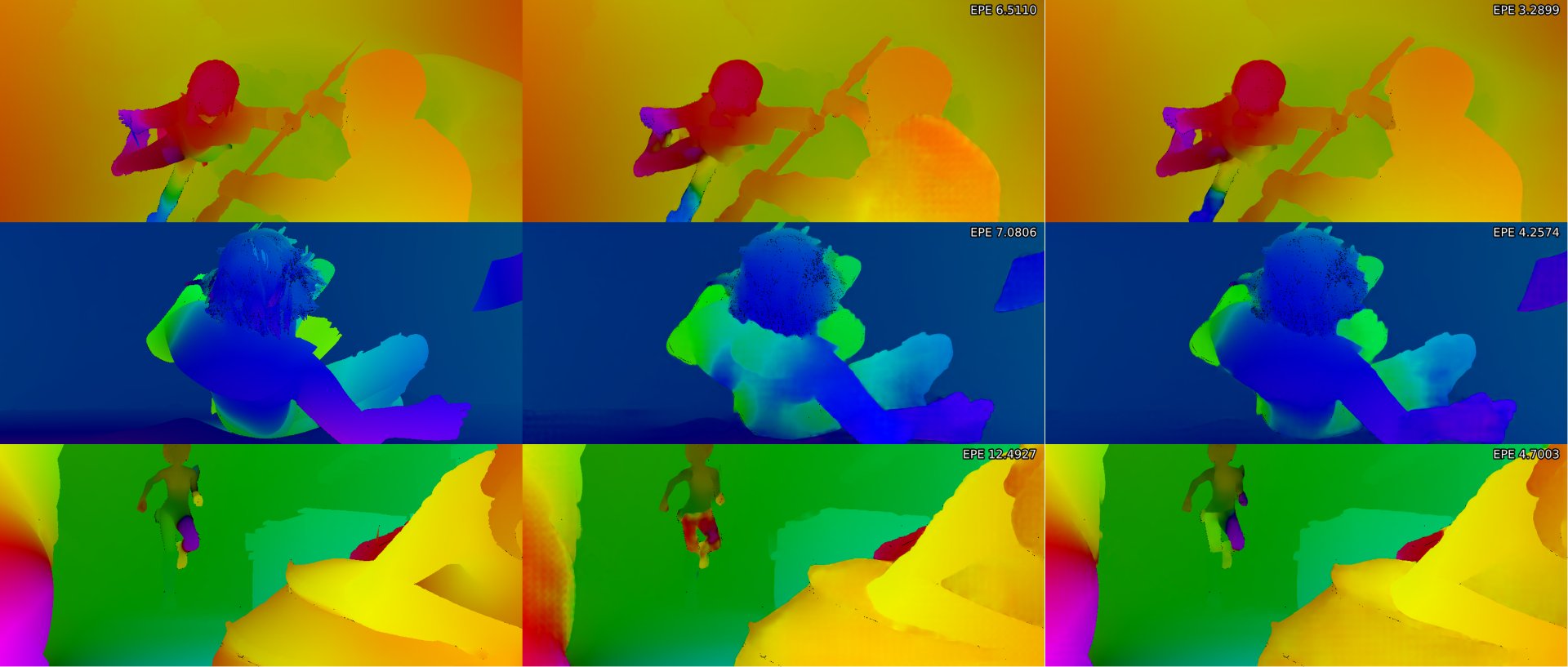}
\caption{Visualisation of the inference of the zero-shot SAGE and WAFT models on Sintel (train) final; from left to right: ground truth, WAFT, SAGE. More results are given in the supplementary material.}
\label{fig:qual}
\end{figure*}

\begin{table}[t]
\centering\scriptsize\setlength{\tabcolsep}{1.9pt}
\renewcommand{\arraystretch}{0.95}%
\begin{tabular}{@{}lccccccc@{}}
\toprule
& \multicolumn{2}{c}{Sintel} & \multicolumn{2}{c}{KITTI} & \multicolumn{3}{c}{Spring} \\
\cmidrule(lr){2-3}\cmidrule(lr){4-5}\cmidrule(lr){6-8}
Method & Clean$\downarrow$ & Final$\downarrow$ & Fl-all$\downarrow$ & Fl-epe$\downarrow$ & EPE$\downarrow$ & 1px$\downarrow$ & Fl$\downarrow$ \\
\midrule
WAFT & 1.28 & 2.56 & 12.9 & 3.49 & 0.18 & 1.02 & 0.43 \\
SAGE & \textbf{0.965} & \textbf{2.141} & 11.366 & \underline{3.237} & 0.185 & 1.317 & \underline{0.419} \\
\addlinespace[2.4pt]
SAGE Laplace & 1.016 & 2.664 & \underline{11.349} & \textbf{3.200} & \underline{0.179} & \textbf{1.014} & 0.439 \\
SAGE learn$\mu$ & 1.021 & 2.411 & 12.065 & 3.353 & 0.183 & 1.154 & 0.421 \\
SAGE $K{=}3$ & \underline{0.973} & \underline{2.379} & 11.414 & 3.239 & \textbf{0.177} & \underline{1.017} & 0.431 \\
SAGE $K{=}7$ & 0.990 & 2.394 & \textbf{11.183} & 3.263 & 0.180 & 1.114 & \textbf{0.407} \\
\bottomrule
\end{tabular}
\caption{Ablation of the design parameters of SAGE: the kernel function, $\mu$ and the value of $K$.}
\label{tab:ablation}
\end{table}

\subsection{Ablation Experiment}
\label{sec:ablations}

\paragraph{K.}SAGE sets $K{=}5$, and \cref{tab:ablation} shows that both $K{=}3$ and $K{=}7$ make Sintel final worse ($2.379$ and $2.394$ vs.\ $2.141$), yet both improve Spring: $K{=}3$ attains the lowest overall EPE ($0.177$), $K{=}7$ the lowest Fl ($0.407$) and the best KITTI Fl-all ($11.183$). The choice of $K$ affects the receptive field size of the correlation feature sampling kernel as well as the GPU memory footprint and computational cost of the entire model. With a small $K$, SAGE samples over a reduced range, improving performance in the small-displacement regions of Sintel clean and naturally reducing both the interfering features entering the correlation feature set and the amount of weighted averaging; the iterative refinement process then receives more accurate feature inputs. This is consistent with the above analysis that SAGE suffers decreased accuracy in the 1-pixel region on Spring sub-val when 
$K{=}5$. Increasing $K$ allows SAGE to sample correlation features over a larger range, better handling challenging regions such as large displacements; at the same time, as noted above, SAGE's feature selection in small-displacement regions introduces less bias, but it incurs increased GPU memory usage and computational cost. In summary, 
$K{=}5$ is the optimal choice after comprehensively considering performance, memory footprint, and computational cost.
\begin{table}[t]
\centering\scriptsize\setlength{\tabcolsep}{1.9pt}
\renewcommand{\arraystretch}{0.95}%
\begin{tabular}{@{}lccccccc@{}}
\toprule
& \multicolumn{2}{c}{Sintel} & \multicolumn{2}{c}{KITTI} & \multicolumn{3}{c}{Spring} \\
\cmidrule(lr){2-3}\cmidrule(lr){4-5}\cmidrule(lr){6-8}
Method & Clean$\downarrow$ & Final$\downarrow$ & Fl-all$\downarrow$ & Fl-epe$\downarrow$ & EPE$\downarrow$ & 1px$\downarrow$ & Fl$\downarrow$ \\
\midrule
DINOv3 WAFT & 1.28 & 2.56 & 12.9 & 3.49 & \textbf{0.18} & \textbf{1.02} & 0.43 \\
DINOv3 SAGE & \textbf{0.97} & \textbf{2.14} & \textbf{11.37} & \textbf{3.24} & 0.19 & 1.32 & \textbf{0.42} \\
\addlinespace[2.4pt]
DAv2 WAFT & \textbf{1.01} & 2.49 & \textbf{10.9} & \textbf{3.28} & 0.21 & 1.45 & 0.39 \\
DAv2 SAGE & 1.24 & \textbf{2.35} & 13.05 & 3.89 & \textbf{0.20} & \textbf{1.15} & \textbf{0.36} \\
\addlinespace[2.4pt]
Twins WAFT & \textbf{1.02} & 2.46 & \textbf{9.9} & \textbf{2.98} & 0.20 & 1.32 & 0.50 \\
Twins SAGE & 1.11 & \textbf{2.52} & 10.89 & 3.00 & \textbf{0.20} & \textbf{1.15} & \textbf{0.46} \\
\bottomrule
\end{tabular}
\caption{For different pretrained input encoders, SAGE improves zero-shot Sintel final over the baseline in every case.}
\label{tab:encoder}
\end{table}
\begin{table*}[t]
\centering\scriptsize\setlength{\tabcolsep}{12.5pt}
\renewcommand{\arraystretch}{0.95}%
\begin{tabular}{@{}lcccccccc@{}}
\toprule
& \multicolumn{2}{c}{Sintel} & \multicolumn{2}{c}{KITTI} & \multicolumn{4}{c}{Spring} \\
\cmidrule(lr){2-3}\cmidrule(lr){4-5}\cmidrule(lr){6-9}
Method & Clean$\downarrow$ & Final$\downarrow$ & All$\downarrow$ & N-Occ$\downarrow$ & 1px$\downarrow$ & EPE$\downarrow$ & Fl$\downarrow$ & WAUC$\uparrow$ \\
\midrule
GMFlow~\cite{gmflow} & 1.74 & 2.90 & 9.32 & 3.80 & 10.355 & 0.945 & 2.952 & 82.337 \\
PWC-Net+~\cite{pwcnet_plus} & 3.45 & 4.60 & 7.72 & 4.91 & -- & -- & -- & -- \\
RAFT~\cite{raft} & 1.61 & 2.86 & 5.10 & 3.07 & 6.790 & 1.476 & 3.198 & 90.920 \\
DIP~\cite{dip} & 1.44 & 2.83 & 4.21 & 2.43 & -- & -- & -- & -- \\
GMFlowNet~\cite{gmflownet} & 1.39 & 2.65 & 4.79 & 2.75 & -- & -- & -- & -- \\
CRAFT~\cite{craft} & 1.45 & 2.42 & 4.79 & 3.02 & -- & -- & -- & -- \\
FlowFormer~\cite{flowformer} & 1.20 & 2.12 & 4.68 & 2.69 & 6.510 & 0.723 & 2.384 & 91.679 \\
GMFlow+~\cite{gmflow_plus} & 1.03 & 2.37 & 4.49 & 2.40 & -- & -- & -- & -- \\
RPKNet~\cite{rpknet} & 1.31 & 2.65 & 4.64 & 2.71 & 4.809 & 0.657 & 1.756 & 92.638 \\
CCMR+~\cite{ccmr} & 1.07 & 2.10 & 3.86 & \textbf{2.07} & -- & -- & -- & -- \\
MatchFlow(G)~\cite{matchflow} & 1.16 & 2.37 & 4.63 & 2.77 & -- & -- & -- & -- \\
FlowFormer++~\cite{flowformer_pp} & 1.07 & \textbf{1.94} & 4.52 & -- & -- & -- & -- & -- \\
SEA-RAFT(L)~\cite{sea_raft} & 1.31 & 2.60 & 4.30 & -- & -- & -- & -- & -- \\
AnyFlow~\cite{anyflow} & 1.23 & 2.44 & 4.41 & 2.69 & -- & -- & -- & -- \\
FlowDiffuser~\cite{flowdiffuser} & 1.02 & 2.03 & 4.17 & 2.82 & -- & -- & -- & -- \\
SAMFlow~\cite{samflow} & 1.00 & 2.08 & 4.49 & -- & -- & -- & -- & -- \\
DPFlow~\cite{dpflow} & 1.04 & \underline{1.97} & \textbf{3.56} & 2.12 & \underline{3.442} & 0.340 & 1.311 & 94.980 \\
SpyNet~\cite{spynet} & 6.64 & 8.36 & 35.07 & 26.71 & 29.963 & 4.162 & 12.866 & 67.150 \\
FlowNet2~\cite{flownet2} & 4.16 & 5.74 & 10.41 & 6.94 & 6.710 & 1.040 & 2.823 & 90.907 \\
\midrule
WAFT & \underline{0.94} & 2.02 & \textbf{3.56} & 2.13 & \textbf{3.182} & \underline{0.325} & \textbf{1.246} & \textbf{95.051} \\
SAGE (ours) & \textbf{0.85} & 2.49 & \underline{3.58} & \underline{2.11} & 3.597 & \textbf{0.306} & \underline{1.264} & \underline{94.556} \\
\bottomrule
\end{tabular}
\caption{We report the performance of SAGE on the public Sintel, KITTI and Spring test benchmarks. SAGE is better than WAFT on Sintel clean, KITTI Non-Occ and Spring EPE. The best entry in each column is in bold and the second best is underlined; ``--'' marks a benchmark on which a method was not reported.}
\label{tab:test}
\end{table*}

\paragraph{Kernel \& Mean.}The Laplace kernel makes both Sintel passes worse ($1.016/2.664$) but improves KITTI Fl-all ($11.349$) and Fl-epe ($3.200$) and gives the lowest Spring 1px error ($1.014$), Near the center of the sampling position, the Laplacian kernel has a sharper and steeper distribution than the Gaussian kernel and filters out more small deviations; however, its tails are heavier, so the features are averaged more strongly, which is unfavorable for estimation in large-displacement occluded regions. Its performance on s40+ degrades (16.078 vs.\ 14.454). 

Learning the centre raises the headline Sintel and KITTI numbers (Fl-all $12.065$), with final s40+ rising from $14.454$ to $16.692$; on Spring its EPE and 1px error improve slightly and Fl barely moves. Fixing the centre is best supported on Sintel and KITTI.

\paragraph{Input Encoders.}We further adapt SAGE to the baseline model with different input-encoder architectures and use zero-shot cross-dataset evaluation to investigate how SAGE behaves on the features produced by these encoders. As shown in \Cref{tab:encoder}, SAGE improves Sintel final with every input encoder: with DINOv3, it improves Sintel clean, Sintel final, and KITTI while degrading slightly on Spring sub-val; with DAv2 and Twins, it improves Spring sub-val but degrades on the other subsets. Overall, these results indicate that the gains from SAGE are consistent, particularly on Sintel final.
Additional metrics for different regions are provided in the supplementary file.

\subsection{Benchmark Submission results}
\label{sec:test}

After longer fine-tuning, SAGE exceeds WAFT by \(9.57\%\) on Sintel test clean (\Cref{tab:test}); on Sintel final scenes, SAGE's weighted sampling is instead less accurate than plain single-point sampling. However, from zero-shot to long-schedule fine-tuning, SAGE achieves a larger improvement than WAFT on Sintel test final (\(62.88\%\) vs.\ \(42.32\%\)); see the supplementary file for details. On KITTI the two are comparable, and on Spring SAGE improves EPE by \(5.85\%\). SAGE improves large-motion regions, and after multi-dataset long-schedule fine-tuning, SAGE improves more than WAFT on difficult benchmarks such as Sintel test final. Given the practical reality of an extreme shortage of large-scale real-world optical flow training data, SAGE's superior zero-shot cross-dataset generalization performance and robust fine-tuning capability are highly valuable.
\section{Conclusion}
\label{sec:conclusion}

In this paper, we study the design space of feature sampling for optical flow estimation, specifically the computation of correlation features between each position in the first frame and the second frame. We propose Anisotropic Gaussian Expectation Sampling (SAGE) as a principled instantiation of this design space, which exposes several key design dimensions---sampling distribution, sampling number, spatial range, and orientation/anisotropy---and makes them position-adaptive. Extensive zero-shot cross-dataset experiments show that occlusion, large displacement, motion boundaries, and weak-texture regions make single-point correlation feature sampling unreliable. SAGE learns a per-position correlation feature sampling distribution and performs weighted sampling of correlation features with different numbers, ranges, and directions at different positions. End-to-end optimization allows these position-adaptive sampling decisions to be learned jointly with the flow network, yielding consistent gains in cross-dataset generalization and suggesting that the benefits stem from the sampling design rather than dataset-specific tuning. Taken together, the proposed design and accompanying analyses help delineate the design space of feature sampling for optical flow, clarifying which sampling dimensions are consequential and how adaptivity along these dimensions contributes to robust matching. We hope this design-space perspective and the empirical findings provide actionable guidance for future work on correlation feature sampling design.

\enlargethispage{6\baselineskip}%
\clearpage
{\setlength{\itemsep}{0pt}\setlength{\parsep}{0pt}%
\bibliographystyle{ieeenat_fullname}\bibliography{main}}
\end{document}